\documentclass{article}

\PassOptionsToPackage{numbers,sort&compress}{natbib}
\usepackage[main, final]{neurips_2026}

\usepackage[utf8]{inputenc}
\usepackage[T1]{fontenc}
\usepackage{hyperref}
\usepackage{url}
\usepackage{booktabs}
\usepackage{amsfonts}
\usepackage{nicefrac}
\usepackage{microtype}
\usepackage{xcolor}

\usepackage{graphicx} 
\usepackage{amsmath}

\usepackage{booktabs}
\usepackage{multirow}
\usepackage[table]{xcolor}
\usepackage{threeparttable}

\definecolor{v2wblue}{RGB}{235,245,255}

\title{From Views to Worlds: Active Exploration \\ over 3D Worlds for Vision-Language Models}
\author{%
  Qijian Tian \\
  Shanghai Jiao Tong University \\
  Shanghai, China \\
  \And
  Jiayu Ying \\
  East China Normal University \\
  Shanghai, China  \\
  \AND
  Ke Fan \\
  Shanghai Jiao Tong University \\
  Shanghai, China \\
  \And
  Lizhuang Ma \\
  Shanghai Jiao Tong University \\
  Shanghai, China \\
  \And
  Xin Tan \\
  East China Normal University \\
  Shanghai, China \\
}

\begin{document}

\maketitle

\begin{abstract}

Vision-language models (VLMs) have achieved remarkable progress in 2D visual understanding tasks, yet the passive view-centric perception paradigm limits their capacity for 3D spatial reasoning. Such reasoning often requires actively acquiring spatial evidence beyond the currently visible views, such as camera poses and cross-view spatial relationships, to infer 3D relationships that are not explicitly represented in 2D images. However, existing methods either lack active exploration or merely provide additional views during inference, without explicitly acquiring such spatial evidence beyond 2D images. In this paper, we propose View-to-World (V2W), an active exploration framework that enables VLMs to actively explore 3D worlds and lifts VLMs from passive view-centric perception to active world-centric spatial reasoning. Given a spatial reasoning task with multi-view images, V2W first reconstructs an explicit 3D world with geometry, camera poses, and language-grounded semantics, and exposes it to VLMs through visual and linguistic interfaces. Through multi-turn interaction with these interfaces, VLMs iteratively acquire spatial evidence from the constructed 3D world, enabling active world-centric spatial reasoning. V2W can be integrated with existing VLMs in a training-free manner and can be further enhanced by optimizing the exploration policy via reinforcement learning. Experiments on spatial mental modeling benchmarks demonstrate that V2W substantially improves VLM baselines in the training-free setting and achieves state-of-the-art performance with the learned exploration policy, highlighting the importance of active exploration for spatial intelligence. Project page: \url{https://fangzhou2000.github.io/projects/v2w}. \par

\end{abstract}

\section{Introduction}
\label{sec:intro}

\begin{figure}[t!]
    \centering
    \includegraphics[width=0.99\textwidth]{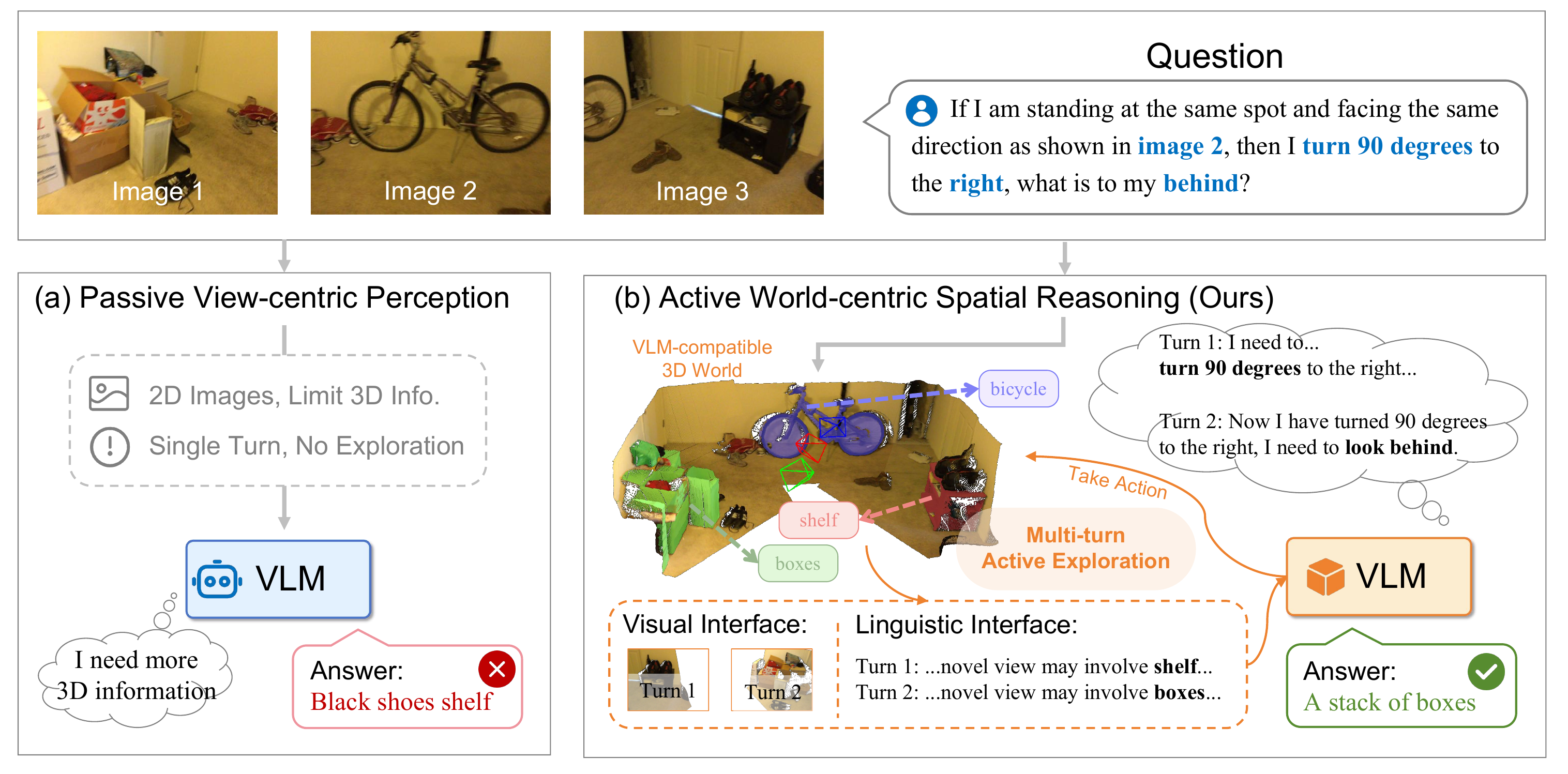}
    \caption{Comparison of passive view-centric perception and active world-centric spatial reasoning}
    \label{fig: teaser}
\end{figure}

Recent advances in vision-language models (VLMs) have demonstrated remarkable capabilities in 2D visual understanding tasks, including image captioning~\cite{personalizing, caparena}, visual question answering~\cite{blip, instructblip, mmmu}, and visual grounding~\cite{kosmos2, qwen2.5vl, qwen3vl}. Building on their strong image-level perception and reasoning capabilities, recent works~\cite{spatialvlm, spatialrgpt} have begun to explore the potential of VLMs for 3D spatial reasoning. However, 2D images are inherently partial observations of the 3D world from specific viewpoints, making view-centric perception fundamentally limited for understanding 3D space. This limitation stems from two key factors. First, critical spatial evidence may be occluded or absent from the currently visible views, which requires active viewpoint changes beyond fixed-view inputs. Second, even with more viewpoints, 2D images alone fail to explicitly represent the cross-view 3D information, such as relative camera poses, cross-view entity correspondences, and object-level 3D spatial relations. Addressing these limitations requires models to actively acquire evidence over a coherent 3D world representation that integrates geometric, visual, and semantic information. \par

However, most existing methods still rely on passively perceiving 2D inputs for spatial reasoning. Some methods attempt to incorporate geometric priors into VLMs, including encoding 3D geometry into the representation space~\cite{llava3d,leovl,video3dllm,gpt4scene,spatialmllm,map2thought}, fusing implicit 3D information at the feature level~\cite{vlm3r,vgllm,geothinker,gapmllm}, and introducing 3D priors through distillation~\cite{3drs,ross3d,seeingthroughimagination} or latent reasoning~\cite{machinementalimagery,lvr,3dthinker}. Although these methods provide VLMs with additional 3D priors, they mainly improve the internal representation capability rather than provide VLMs with an explicitly interactive and actively explorable 3D world, and therefore still belong to the passive perception from fixed-view input. More recent methods~\cite{think3d, mindjourney, spatialdreamer} have further explored active perception by providing VLMs with additional 2D observations. Nevertheless, these methods remain largely view-centric, as additional observations are still presented as 2D views and lack cross-view 3D information needed for spatial reasoning. Such 3D information is not explicitly represented in 2D views, but is grounded in the underlying 3D world. We refer to reasoning over such information as world-centric spatial reasoning. Therefore, advancing VLMs toward
3D spatial reasoning requires moving from passive view-centric perception to active world-centric spatial reasoning, as shown in Fig.~\ref{fig: teaser}. \par

In this paper, we propose View-to-World (V2W), a model-agnostic framework that enables VLMs to actively explore over an explicit 3D world representation. Given multi-view images, V2W first recovers scene geometry and camera poses with off-the-shelf 3D foundation models and incorporates language-grounded semantics to build a geometrically consistent, semantically aligned, and language-queryable 3D world. We denote it as a VLM-compatible 3D world, since the geometry, camera poses, and semantic evidence are exposed to VLMs through images and natural language rather than raw 3D representations. We design two interfaces to achieve the exposure: a visual interface that provides 2D observations from newly explored viewpoints and a linguistic interface that converts geometric and semantic information from the 3D world into explicit natural-language cues. V2W enables VLMs to actively explore the constructed 3D world through visual and linguistic interfaces and progressively acquire and integrate spatial evidence via multi-turn interaction, thereby performing spatial reasoning in an active world-centric manner. Experiments on spatial mental modeling benchmarks show that V2W improves existing VLMs in a training-free manner, revealing their inherent capability for active exploration. However, we also find that smaller VLMs often struggle to perform effective active exploration, exhibiting insufficient or uninformative exploration. To this end, we further propose to train the exploration policy through reinforcement learning, thereby improving the active exploration capability. With the learned exploration policy, V2W achieves state-of-the-art performance, substantially outperforming both baseline methods and proprietary commercial models (e.g., surpassing GPT-4o by 48.5 percentage points on MindCube-Tiny), demonstrating the importance of active exploration capability for spatial reasoning. Our contributions are summarized as follows:

\begin{itemize}
    \item We construct a VLM-compatible 3D world representation that is geometrically consistent, semantically aligned, and language-queryable, providing a 3D environment that lifts VLMs from view-centric perception to world-centric spatial reasoning. \par
    \item We propose V2W, an active exploration framework that enables VLMs to interact with the constructed 3D world through visual and linguistic interfaces and progressively acquire world-grounded evidence via multi-turn exploration. \par
    \item We introduce reinforcement learning to train the exploration policy of V2W, enabling more effective evidence acquisition from the constructed 3D world. \par
    \item Experiments on spatial mental modeling benchmarks demonstrate that V2W substantially improves VLM baselines in the training-free setting and achieves state-of-the-art performance with the learned exploration policy. \par
\end{itemize}

\section{Related Work}

\subsection{Vision-Language Models for Spatial Reasoning}

Spatial reasoning is essential for intelligent perception and interaction. It allows agents to represent, organize, and transform objects and their spatial relationships in 3D environments~\cite{mentalmodelsincognitivescience, mental, spatialreasoning, exemplar2vqa}. However, recent studies show that existing VLMs still struggle with spatial reasoning~\cite{scanrefer, scanqa, scan2cap, multi3drefer, 3dbench, sat, vsibench, beacon3d, spar, spatialbench, mindcube}. Prior efforts have attempted to improve the capability of VLMs mainly in two ways. The first line of work injects 3D priors into VLM representations. These methods encode 3D geometry into the representation space~\cite{llava3d, leovl, video3dllm, gpt4scene, spatialmllm, map2thought}, fuse implicit 3D information at the feature level~\cite{vlm3r,vgllm, geothinker, gapmllm}, or introduce 3D knowledge through distillation~\cite{3drs, ross3d, seeingthroughimagination} or latent reasoning~\cite{machinementalimagery, lvr, 3dthinker}. While effective, they often require large-scale aligned data and costly training, and they do not provide explicit geometric constraints. The second line of work provides 3D information externally in the VLM context~\cite{pyspatial, mssr, gca, think3d, pigeon}, which offers a more direct and model-agnostic way to improve 3D spatial awareness. However, a key challenge is how to help VLMs understand and use this external 3D information effectively. In this work, we also provide 3D information externally, but we construct a VLM-compatible 3D world that supports active exploration, allowing VLMs to make better use of 3D information. \par

\subsection{3D Foundation Models and Scene Understanding}

Recent feed-forward reconstruction models~\cite{dust3r, mast3r, drivingforward, vggt, da3, pi3} have made rapid progress in 3D modeling. These models can recover 3D geometry from multi-view images and accurately estimate camera parameters and other 3D information. Their strong generalization ability further positions them as 3D foundation models. In parallel, visual foundation models, such as the SAM series~\cite{sam, sam2, sam3}, have demonstrated strong capabilities in scene parsing and understanding. Recent studies~\cite{iggt, ff3r, uniforward, fleg, groupforward} further integrate 3D and visual foundation models to support semantic 3D scene understanding. While the above work mainly focuses on 3D representation and perception-oriented tasks, our work investigates how VLMs can benefit from such foundation models to improve spatial reasoning capability. \par

\subsection{Vision-Language Models with Active Interaction}

Recent VLMs have advanced from direct reasoning over given inputs to interactive reasoning~\cite{react, mmreact, chameleon}. Representative methods allow models to interact with external resources during inference through multi-turn reasoning, including tool-augmented VLMs~\cite{hugginggpt, deepeyes, vtoolr1, vilasr, openthinkimg} and visual programming methods~\cite{visprog, vipergpt, vadar, pyspatial}. Such interaction reflects the active exploration capability of VLMs~\cite{activeo3, actiview, sieve}. This capability is particularly important for spatial reasoning, since understanding 3D space often requires integrating observations from different viewpoints and reasoning about regions beyond the current field of view. Recent works have introduced active exploration into spatial reasoning through novel-view rendering~\cite{think3d} or generation~\cite{mindjourney, spatialdreamer}. However, these methods mainly provide additional 2D views to VLMs and therefore remain view-centric, without explicitly exposing 3D information beyond 2D images. Another work~\cite{pyspatial} performs active exploration through code generation, but this paradigm restricts multi-turn exploration and relies heavily on the code-generation ability of VLMs. In contrast, V2W constructs a VLM-compatible 3D world with visual and linguistic interfaces, enabling VLMs to gather spatial evidence through multi-turn interaction and shift from passive view-centric perception to active world-centric exploration. \par

\label{sec:related}

\section{Methods}\label{sec:methods}

\begin{figure}[t!]
    \centering
    \includegraphics[width=0.99\textwidth]{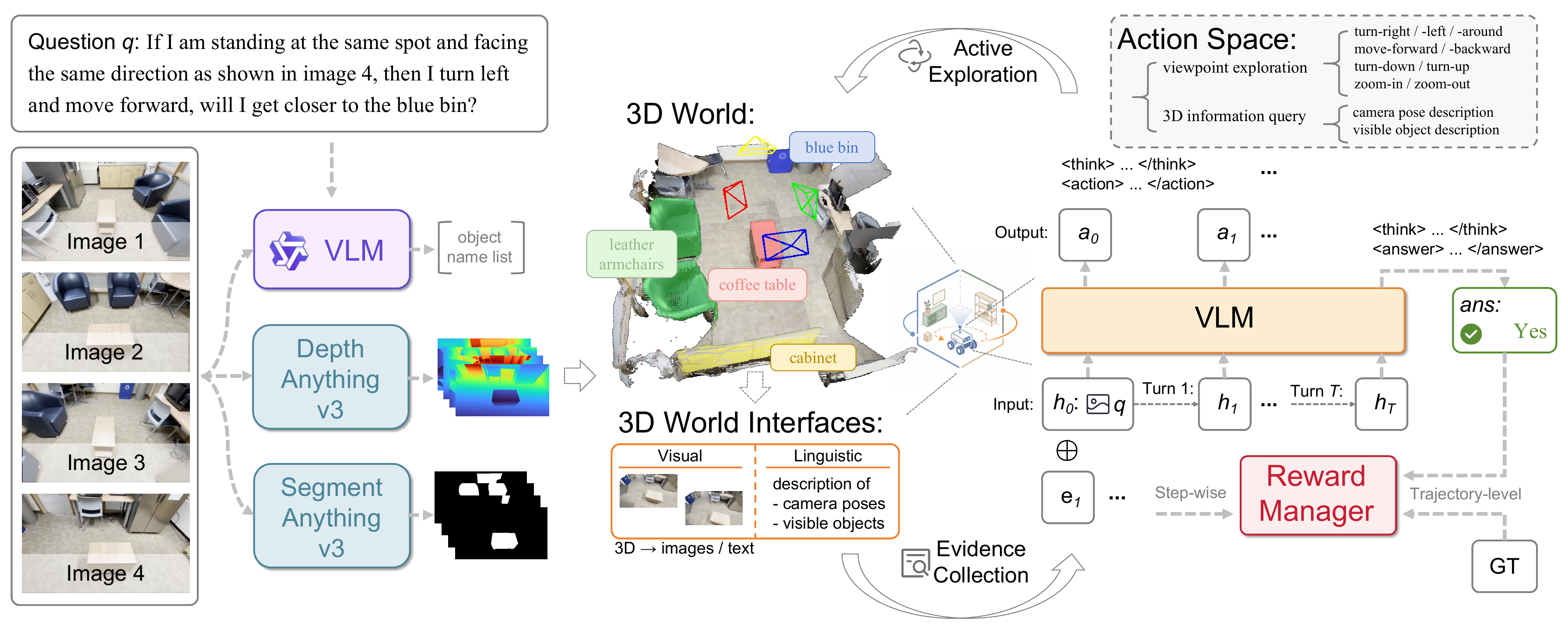}
    \caption{Overview of the View-to-World (V2W) framework. V2W constructs a VLM-compatible 3D world from multi-view images. The 3D information is exposed through visual and linguistic interfaces, enabling the VLM to actively explore the world via multi-turn interaction, and the exploration policy can be further optimized via reinforcement learning.}
    \label{fig: main}
\end{figure}

\subsection{Overview}

Given multi-view images $\mathcal{I}=\{I_i\}_{i=1}^{N}$ and a spatial reasoning question $q$, V2W aims to produce the answer by enabling a VLM to actively explore an explicit 3D world. As shown in Fig.~\ref{fig: main}, V2W first constructs a VLM-compatible 3D world $\mathcal{W}$ by reconstructing the 3D scene and lifting question-relevant object semantics into 3D space (Sec.~\ref{3d world}). Since VLMs only take images and text as input, $\mathcal{W}$ is exposed through two interfaces: a visual interface that renders novel-view images, and a linguistic interface that provides textual cues about visible objects and relative camera poses. During inference, V2W formulates active exploration as a multi-turn interaction via function call (Sec.~\ref{exploration}). We further train the exploration policy with supervised fine-tuning and reinforcement learning (Sec.~\ref{rl}), which strengthens the VLM's ability to actively explore the 3D world. \par

\subsection{VLM-compatible 3D World Construction}\label{3d world}

Active exploration requires an explicit 3D world that serves as an interactive environment for existing VLMs. Since VLMs typically take images and natural language as inputs, raw 3D representations such as point clouds, meshes, or 3D Gaussian Splatting cannot be directly used. Meanwhile, spatial reasoning often requires both geometry and language-grounded semantics. Therefore, we construct a VLM-compatible 3D world that integrates scene geometry, camera poses, and object-level semantics, and exposes them to VLMs through visual and linguistic interfaces. These interfaces convert explicit 3D information into rendered observations and natural-language cues, allowing VLMs to acquire spatial evidence without modifying their architectures. \par

\paragraph{Geometric Reconstruction and Semantic Lifting.} 

We first use an off-the-shelf 3D foundation model $\Phi_{\mathrm{3D}}$~\cite{da3} to estimate the geometry and camera parameters from the input images $\mathcal{I} =\{I_i\}_{i=1}^{N}$,
\begin{equation}
    \{\mathbf{D}_i, \mathbf{E}_i, \mathbf{K}_i\}_{i=1}^{N}
    = \Phi_{\mathrm{3D}}(\mathcal{I}),
\end{equation}
where $\mathbf{D}_i \in \mathbb{R}^{H \times W}$ is the predicted depth map, $\mathbf{E}_i \in \mathrm{SE}(3)$ is the camera-to-world pose, and $\mathbf{K}_i$ is the camera intrinsic. We then obtain the point map by back-projecting each pixel into 3D space:
\begin{equation}
    \mathbf{P}_i(\mathbf{u})
    =
    \pi^{-1}(\mathbf{u}, \mathbf{D}_i(\mathbf{u}); \mathbf{K}_i, \mathbf{E}_i)
    =
    \mathbf{R}_i
    \left(
    \mathbf{D}_i(\mathbf{u}) \mathbf{K}_i^{-1} \tilde{\mathbf{u}}
    \right)
    +
    \mathbf{t}_i,
\end{equation}
where $\mathbf{u}=(u,v)$ is a pixel coordinate, $\tilde{\mathbf{u}}=(u,v,1)^\top$ is its homogeneous form, and $\mathbf{E}_i=[\mathbf{R}_i|\mathbf{t}_i]$. The resulting point map $\mathbf{P}_i \in \mathbb{R}^{H \times W \times 3}$ assigns each pixel in $I_i$ a 3D coordinate in the global coordinate system. The point maps provide the scene geometry, while the camera poses establish the spatial relationships among input views. To incorporate semantics, we use a lightweight VLM, Qwen2.5-VL-7B~\cite{qwen2.5vl}, to extract question-relevant object names from the input images $\mathcal{I}$ and the question $q$, and then use these objects as text prompts for SAM 3~\cite{sam3} to obtain their 2D masks in the input views. The masks are lifted to 3D by associating masked pixels with their corresponding 3D points in the point maps. For an object $o$, its 3D semantic point set is defined as
\begin{equation}
    \mathcal{X}_{o} =
    \left\{
        \mathbf{P}_i(\mathbf{u}) \mid
        \mathbf{M}_{i,o}(\mathbf{u})=1,\ 
        i=1,\dots, N
    \right\},
\end{equation}
where $\mathbf{M}_{i,o}$ is the mask of object $o$ in view $I_i$, and $\mathbf{u}$ is a pixel location. This step lifts language-grounded object semantics onto the reconstructed geometry, yielding a 3D representation with object-level semantic information. We additionally obtain 3D floor points through the same mask-lifting process, and align the reconstructed 3D world with the real scene by estimating the gravity direction from the floor plane using RANSAC~\cite{ransac}. This allows the subsequent action space to be defined on the horizontal plane with intuitive directions such as forward, backward, left, and right.\par

\paragraph{Visual and Linguistic Interfaces.}

The constructed 3D representation contains both geometry and object-level semantics, whereas existing VLMs typically take only images and text as input. To bridge this modality gap, we expose the 3D world through two VLM-compatible interfaces: the visual and linguistic interfaces. The visual interface returns rendered images from queried camera poses, while the linguistic interface returns textual cues derived from object-level semantics and spatial relations. \par

Given a queried camera pose, the visual interface renders the corresponding novel-view image as the visual observation for the VLM. To support object-level textual cues in the linguistic interface, we internally render object masks in the queried view. Specifically, for a target pose $\mathbf{E}_{\tau}$, the mask of object $o$ is obtained by projecting its associated 3D points into the target view:
\begin{equation}
    \mathcal{P}_{o}
    =
    \left\{
    \mathbf{P}_{i}(\mathbf{u})
    \mid
    \mathbf{M}_{i,o}(\mathbf{u})=1,
    i=1,\ldots, N
    \right\},
    \qquad
    \hat{\mathbf{M}}_{\tau,o}
    =
    \Pi_{\tau}(\mathcal{P}_{o}).
\end{equation}
where $\mathcal{P}_{o}$ denotes the set of 3D points lifted from the masks of object $o$ across all input views, and $\Pi_{\tau}(\cdot)$ projects these points into the target view to obtain the internal mask $\hat{\mathbf{M}}_{\tau,o}$. The projected mask is used to identify visible objects, which are then returned as textual cues by the linguistic interface. The linguistic interface also describes spatial relations between camera poses. Given two camera poses and the gravity direction, we project the camera centers onto the horizontal plane to describe their relative translation as forward, backward, left, right, or diagonal, and compare their orientations to describe the relative rotation as turning left or right. Together, the visual and linguistic interfaces provide the VLM with image observations and textual spatial cues through its native input modalities. \par

\subsection{Active Exploration via Function Calling}\label{exploration}

The visual and linguistic interfaces of the constructed VLM-compatible 3D world are formulated as callable functions, enabling V2W to perform active exploration as a structured tool-calling process. Each function corresponds to a predefined exploration action. During inference, the VLM calls a function with specific arguments, receives the returned images or text from the visual or linguistic interface, and then decides whether to continue exploration or produce the final answer. \par

\paragraph{Exploration Action Space.}

V2W defines an exploration action space with two types of actions: viewpoint exploration and 3D information query, as shown in Fig.~\ref{fig: main}. Viewpoint exploration actions change the current camera pose to acquire additional visual evidence from novel views, including horizontal rotations around the gravity direction, i.e., turn left, turn right, and turn around; translations on the horizontal plane, i.e., move forward and move backward; and view-centered adjustments, i.e., zoom in, zoom out, turn up, and turn down. After each viewpoint exploration action, the 3D world returns a rendered novel-view image through the visual interface. 3D information query actions provide textual spatial cues from the 3D world through the linguistic interface. These cues include visible object descriptions, which summarize the object categories observed from the explored viewpoint, and relative camera pose descriptions, which return the relative translation and rotation between two existing views in the gravity-aligned coordinate system. \par

\paragraph{Multi-turn Interaction.}

During inference, V2W formulates exploration as a multi-turn interaction between the VLM and the constructed 3D world. Let $\mathcal{W}$ denote the VLM-compatible 3D world, $h_t$ denote the cumulative interaction history up to turn $t$, including the original question, previous function calls, and the accumulated evidence, and $\pi_\theta$ denote the VLM policy parameterized by $\theta$, which conditions on the interaction history to generate exploration actions and the final answer. Starting from $h_0=(\mathcal{I},q)$, the VLM iteratively performs an exploration action, receives world-grounded evidence, and updates the interaction context:
\begin{equation}
a_t = \pi_{\theta}^{\mathrm{act}}(h_t),
\qquad
e_{t+1} = \mathcal{W}(a_t),
\qquad
h_{t+1} = h_t \oplus (a_t, e_{t+1}),
\end{equation}
where $a_t$ denotes the exploration action selected by the VLM, $\pi_{\theta}^{\mathrm{act}}$ denotes using the policy to generate the next exploration action, $e_{t+1}$ denotes the returned visual or linguistic evidence, and $\oplus$ denotes appending the new action and evidence to the context. This recursive process allows the VLM to progressively accumulate world-grounded spatial evidence. When sufficient evidence has been collected at turn $T$, the VLM produces the final answer based on the interaction history:
\begin{equation}
\mathrm{ans} = \pi_{\theta}^{\mathrm{ans}}(h_T).
\end{equation}
where $\pi_{\theta}^{\mathrm{ans}}$ denotes using the same policy to generate the final answer from interaction history $h_T$.
\par

\subsection{Exploration Policy Learning}\label{rl}

Although V2W can be applied to existing VLMs without training, smaller VLMs often struggle with active exploration. They may terminate prematurely, repeat uninformative actions, or select actions weakly related to the question. This suggests that effective exploration also depends on the VLM's exploration policy. We therefore optimize this policy using supervised fine-tuning (SFT) and reinforcement learning (RL). \par

\paragraph{Supervised Fine-tuning.} To learn an effective exploration policy, we first perform supervised fine-tuning on demonstration trajectories before reinforcement learning. This stage teaches the model the function-calling format, valid exploration actions, and how to reason over accumulated evidence. We use Qwen3-VL-235B~\cite{qwen3vl} as the teacher model to generate trajectories. Instead of generating a full trajectory at once, the teacher interacts with the VLM-compatible 3D world turn by turn. At each turn, it predicts the next reasoning step and function call based on the current interaction history, receives the returned visual or linguistic evidence, and appends the new step to the trajectory. After the interaction ends, we concatenate the reasoning steps, function calls, returned evidence, and final answer into a complete trajectory. Each trajectory follows the format:
\texttt{<think>} reasoning \texttt{</think>}
\texttt{<action>} function call \texttt{</action>}
$\cdots$
\texttt{<think>} reasoning \texttt{</think>}
\texttt{<answer>} final answer \texttt{</answer>}.
Here, each \texttt{<think>} segment contains intermediate reasoning, each \texttt{<action>} segment specifies a function call with its arguments, and the final \texttt{<answer>} segment gives the answer. We filter out invalid trajectories with malformed function calls, failed executions, or incorrect answers, and fine-tune the VLM on the filtered data using a standard next-token prediction objective. \par

\paragraph{Reinforcement Learning.} 
After supervised fine-tuning, we optimize the exploration policy with reinforcement learning. We adopt Group Relative Policy Optimization (GRPO)~\cite{grpo} as the policy optimization algorithm. For reward design, we first introduce three trajectory-level rewards, including a format reward that enforces valid function calls and answer formats, an answer reward that evaluates whether the final prediction is correct, and an exploration reward that checks whether the trajectory contains valid exploration actions. \par

Since trajectory-level rewards provide only coarse supervision, we further introduce step-wise rewards to guide intermediate exploration decisions. For each training sample, we use Qwen3-VL-235B~\cite{qwen3vl} to pre-generate question-conditioned preferred exploration actions. During training, the policy receives a step-wise matching reward when its selected action is consistent with the preferred action at each interaction step. We also apply a step-wise action penalty for failed function calls and uninformative actions, such as actions that lead to entirely empty new views from the visual interface. The overall reward for a trajectory $\tau$ is defined as
\begin{equation} 
R(\tau) = \underbrace{ r_{\mathrm{fmt}} + r_{\mathrm{ans}} + r_{\mathrm{explore}} }_{\text{trajectory-level}} + \underbrace{ \lambda r_{\mathrm{match}} - \beta p_{\mathrm{act}} }_{\text{step-wise}} 
\end{equation}
where $r_{\mathrm{match}}$ denotes the step-wise action matching reward, $p_{\mathrm{act}}$ denotes the step-wise action penalty, and $\lambda$ and $\beta$ are weighting coefficients. \par

\section{Experiments}
\label{sec:experiments}

\subsection{Experimental Settings}

\paragraph{Benchmarks.} 

Spatial mental modeling refers to internal representations of unseen space that support reasoning about layout, perspective, and motion~\cite{mental, mentalmodelsincognitivescience}. It is closely aligned with our goal of evaluating whether VLMs can perform human-like spatial reasoning from limited observations. We therefore evaluate our framework on two related benchmarks: (1) \textbf{MindCube} contains 21,154 questions over 3,268 images and focuses on inferring spatial layouts from limited observations, particularly through what-if questions that require mental simulation. Its training set contains 10K samples, and the commonly used MindCube-Tiny evaluation set includes 1,050 questions, comprising 600 Among, 200 Rotation, and 250 Around questions. Following prior work~\cite{mindcube, pyspatial, 3dthinker, spatialdreamer}, we primarily evaluate our method on MindCube-Tiny, and the overall accuracy is weighted by the number of questions in each category. (2) \textbf{SAT} similarly evaluates human-like spatial reasoning from sparse visual observations across diverse categories, including Egocentric Movement, Object Movement, Allocentric Perspective, Goal Aiming, and Action Consequence. It also includes questions that require mental simulation. Its training set contains 175K questions across 22K procedurally generated scenes, and the SAT-val evaluation set contains 4,000 questions. We evaluate our method on SAT-val, and the overall accuracy is averaged equally across the five categories, following the original SAT protocol~\cite{sat}. \par  

\paragraph{Baselines.}

Our framework can be applied to existing VLMs in a training-free manner and also supports further optimization of the VLM exploration policy via RL. We therefore evaluate our framework under both training-free and training-based experimental settings. The baselines are grouped into four categories: (1) generalist open-source VLMs, including Qwen series: Qwen2.5-VL-7B-Instruct~\cite{qwen2.5vl}, Qwen3-VL-8B-Instruct~\cite{qwen3vl}, Qwen3-VL-235B-A22B-Instruct~\cite{qwen3vl}, and the latest Qwen3.5-9B~\cite{qwen3.5}; (2) proprietary commercial models, e.g. GPT-4o~\cite{gpt4o} and GPT-4.1 series~\cite{gpt41}; (3) training-free spatial reasoning methods, e.g. pySpatial~\cite{pyspatial}; and (4) specialized spatial reasoning VLMs, e.g. 3DThinker~\cite{3dthinker} and SpatialDreamer~\cite{spatialdreamer}, representing training-based methods.

\paragraph{Implementation Details.}

We use off-the-shelf foundation models, including 3D foundation model Depth Anything 3~\cite{da3} and the segmentation model SAM 3~\cite{sam3}, to construct the VLM-compatible 3D world. During exploration, we set the maximum number of interaction turns to 4. For the training-free setting, all evaluations are conducted on a single H200 GPU. Since our framework is model-agnostic, we adapt V2W with different base models to align with the underlying VLMs used by different baselines. For the training-based setting, we train our method separately on the training set of each benchmark using 8 H200 GPUs. To ensure consistency with the main comparison methods~\cite{3dthinker, spatialdreamer}, we also adopt Qwen2.5-VL-7B-Instruct~\cite{qwen2.5vl} as the base model. In the SFT stage, we fine-tune all model parameters for one epoch using a learning rate of $1.0 \times 10^{-5}$, a cosine learning-rate schedule with a 0.1 warmup ratio, and a batch size of 64. In the RL stage, we implement the exploration policy training with the verl~\cite{verl} framework, using a learning rate of $1.0 \times 10^{-6}$ and a batch size of 128. \par

\begin{table}[t]
\centering
\footnotesize
\caption{Quantitative results on MindCube-Tiny under training-free and training-based settings.}
\begin{threeparttable}
\label{tab: mindcube}
\begin{tabular}{lccccc}
\toprule
Method & Setting & \textbf{Overall} & Among & Rotation & Around \\
\midrule
Qwen2.5-VL-7B~\cite{qwen2.5vl}& & 34.5 & 32.7 & 35.0 & 38.4 \\
Qwen2.5-VL-7B~\cite{qwen2.5vl} w CoT &  & 29.1 & 27.7 & 31.5 & 30.8 \\
\rowcolor{v2wblue} V2W (Qwen2.5-VL-7B) &  & 43.8 & 43.7 & 35.0 & 51.2 \\
Qwen3-VL-8B~\cite{qwen3vl} &  & 30.6 & 29.2 & 31.5 & 33.2 \\
Qwen3-VL-8B~\cite{qwen3vl} w CoT &  & 38.9 & 33.7 & 40.0 & 50.4 \\
\rowcolor{v2wblue} V2W (Qwen3-VL-8B) &  & 50.2 & 44.8 & 38.5 & 72.4 \\
Qwen3-VL-235B~\cite{qwen3vl} &  & 42.3 & 38.8 & 42.5 & 50.4 \\
Qwen3-VL-235B~\cite{qwen3vl} w CoT &  & 45.4 & 41.0 & 49.5 & 52.8 \\
\rowcolor{v2wblue} V2W (Qwen3-VL-235B) &  & 59.6 & 60.2 & 49.5 & 66.4 \\
Qwen3.5-9B~\cite{qwen3.5} &  & 33.7 & 26.5 & 33.5 & 51.2 \\
Qwen3.5-9B~\cite{qwen3.5} w CoT &  & 45.9 & 38.7 & 49.0 & 60.8 \\
\rowcolor{v2wblue} V2W (Qwen3.5-9B) &  & 55.4 & 47.7 & 65.0 & 66.4 \\
GPT-4o~\cite{gpt4o} &  & 42.3 & 43.0 & 35.0 & 46.4 \\
GPT-4.1~\cite{gpt41} &  & 44.7 & 45.3 & 35.5 & 50.4 \\
GPT-4.1-mini~\cite{gpt41} &  & 43.3 & 45.0 & 36.0 & 44.8 \\
ViperGPT~\cite{vipergpt} &  & 37.0 & 41.0 & 20.5 & 40.4 \\
VADAR~\cite{vadar} &  & 40.8 & 40.7 & 33.5 & 46.8 \\
pySpatial (GPT-4.1-mini)~\cite{pyspatial} &  & 58.2 & 62.0 & 37.5 & 65.6 \\
\rowcolor{v2wblue} V2W (GPT-4.1-mini) & \multirow{-19}{*}{Training-free} & \textbf{61.7} & 61.3 & 47.0 & 74.4 \\
\midrule
SpaceMantis~\cite{spatialvlm} &  & 26.3 & 20.5 & 35.0 & 33.2 \\
Spatial-MLLM~\cite{spatialmllm} &  & 33.4 & 30.5 & 39.0 & 36.0 \\
Space-Qwen~\cite{spatialvlm} &  & 36.9 & 36.3 & 32.5 & 41.6 \\
RoboBrain~\cite{robobrain} &  & 44.0 & 45.2 & 36.0 & 47.6 \\
3DThinker (Qwen2.5-VL-7B)~\cite{3dthinker} &  & 76.0 & 83.0 & 55.0 & 76.0 \\
SpatialDreamer (Qwen2.5-VL-7B)~\cite{spatialdreamer}$^{\dagger}$ &  & 84.9 & 78.2 & 87.5 & 93.5 \\
\rowcolor{v2wblue} V2W (Qwen2.5-VL-7B) & \multirow{-8}{*}{Training-based}  & \textbf{90.8} & 91.3 & 87.5 & 92.0 \\
\bottomrule
\end{tabular}
\begin{tablenotes}[flushleft]
\scriptsize
\item[$\dagger$] Results are taken from the corresponding original paper.
\end{tablenotes}
\end{threeparttable}
\end{table}

\subsection{Quantitative Results}

We evaluate V2W under both training-free and training-based settings. First, in the training-free setting, we assess whether V2W can serve as a plug-and-play framework to directly enhance the spatial reasoning ability of existing VLMs without additional training. Second, in the training-based setting, we further investigate whether a learned exploration policy can lead to stronger spatial reasoning capabilities. Specifically, we evaluate training-free methods on MindCube-Tiny, and evaluate training-based methods on both MindCube-Tiny and SAT-val. \par

\paragraph{Training-free Evaluation} As shown in Table~\ref{tab: mindcube}, V2W substantially improves the overall performance of different base VLMs on MindCube-Tiny without additional training. V2W improves the overall accuracy of Qwen2.5-VL-7B, Qwen3-VL-8B, Qwen3-VL-235B, and GPT-4.1-mini by 9.3, 19.6, 17.3, and 18.4 percentage points, demonstrating its ability to enhance existing VLMs across different model families and scales. To verify that these gains do not merely arise from eliciting explicit textual reasoning, we further compare V2W with chain-of-thought (CoT) prompting, which encourages the model to perform free-form textual reasoning before producing the final answer. V2W also substantially outperforms CoT under the same base models, suggesting that the improvements are primarily driven by active exploration over explicit 3D world representations. In addition, V2W outperforms other training-free spatial reasoning methods, such as pySpatial, a visual-programming-based approach. Compared with visual programming pipelines, V2W enables multi-turn active exploration via lightweight function calls, thereby avoiding potential failures arising from code generation and execution. \par

\paragraph{Training-based Evaluation} Although training-free V2W already improves the overall performance of existing VLMs, its gains remain limited on more challenging question types such as Rotation, which often involve complex what-if hypotheses and require multi-step exploration to identify critical spatial evidence. This suggests that existing VLMs still lack effective exploration strategies for complex spatial reasoning. To address this limitation, we further optimize the exploration policy via reinforcement learning, enabling VLMs to better conduct multi-step exploration and aggregate spatial evidence. As shown in Table~\ref{tab: mindcube}, training-based V2W achieves 90.8\% overall accuracy on MindCube-Tiny, outperforming existing specialized spatial reasoning models. Table~\ref{tab: sat_val} further shows that V2W also surpasses competing methods on SAT-val, demonstrating that RL-based exploration policy training effectively enhances the active exploration capability of VLMs. \par

\begin{table}[t]
\centering
\footnotesize
\caption{Quantitative results on SAT-val under the training-based setting.}
\label{tab: sat_val}
\begin{tabular}{lcccccc}
\toprule
Method & \textbf{Overall} & EgoM & ObjM & EgoAct & GoalAim & Pers \\
\midrule
GPT-4V~\cite{gpt4v} & 44.8 & 54.7 & 32.7 & 52.0 & 50.5 & 34.2 \\
GPT-4o~\cite{gpt4o} & 63.1 & 64.7 & 86.8 & 51.9 & 68.7 & 43.4 \\
GPT-4.1~\cite{gpt41} & 68.4 & 75.3 & 89.0 & 57.8 & 78.3 & 41.5 \\
o1~\cite{o1} & 74.0 & 78.0 & 85.9 & 65.4 & 86.0 & 54.6 \\
RoboPoint-13B~\cite{robopoint} & 53.3 & 50.2 & 69.4 & 48.8 & 72.6 & 25.5 \\
LLaVA-1.5-13B~\cite{llava1.5} & 51.1 & 46.6 & 73.8 & 49.7 & 45.6 & 39.9 \\
LLaVA-Video-7B~\cite{llavavideo} & 57.4 & 56.4 & 82.7 & 48.0 & 52.9 & 47.1 \\
SAT (LLaVA-1.5-13B)~\cite{sat} & 87.7 & 61.7 & 90.2 & 91.4 & 96.8 & 98.5 \\
SAT (LLaVA-Video-7B)~\cite{sat} & 78.0 & 79.6 & 80.4 & 85.3 & 56.4 & 88.4 \\
SpatialDreamer (Qwen2.5-VL-7B)~\cite{spatialdreamer} & 93.0 & 95.4 & 93.1 & 91.8 & 94.9 & 89.8 \\
\rowcolor{v2wblue} V2W (Qwen2.5-VL-7B) & \textbf{95.5} & 99.8 & 90.3 & 90.5 & 97.8 & 99.2 \\
\bottomrule
\end{tabular}
\end{table}

\begin{figure}[t]
    \centering
    \includegraphics[width=0.99\textwidth]{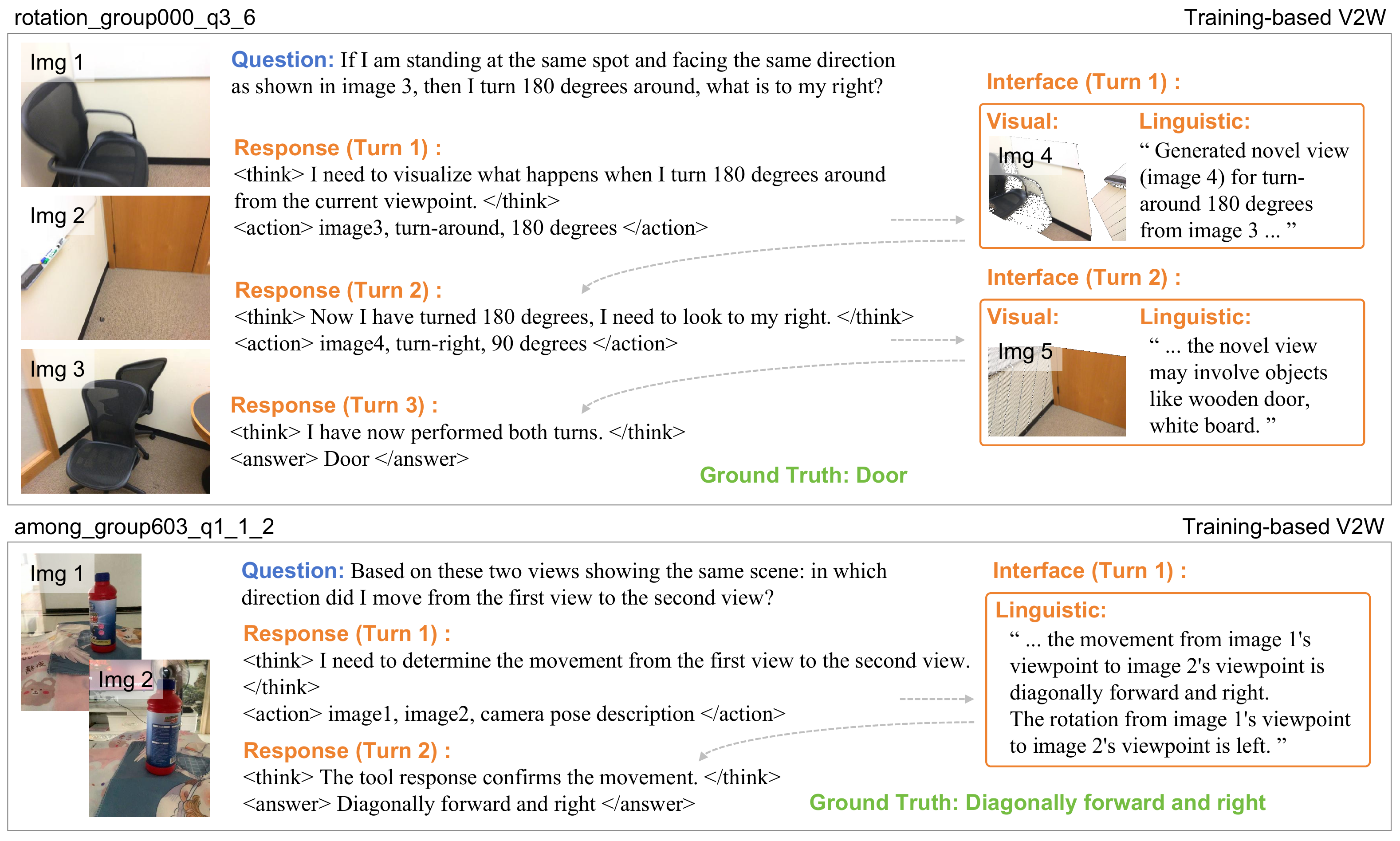}
    \caption{Qualitative results on MindCube-Tiny.}
    \label{fig: case}
\end{figure}

\subsection{Qualitative Analysis}

Fig.~\ref{fig: case} presents the spatial reasoning results of training-based V2W on MindCube-Tiny. The first case illustrates how V2W actively explores the 3D world and collects spatial evidence through both visual and linguistic interfaces over multiple interaction turns. The second case demonstrates that, through our linguistic interface, V2W can actively acquire 3D information that is not explicitly available in the original 2D images, thereby facilitating more effective spatial reasoning. More qualitative results and analyses are provided in the appendix. \par

\subsection{Ablation Study}

\begin{table}[t]
\centering
\footnotesize
\caption{Ablation studies on MindCube-Tiny.}
\label{tab: ablation}
\begin{tabular}{c@{\quad}c@{\quad}cccccccc}
\toprule
\multirow{2}{*}{\#} & \multirow{2}{*}{Setting} & \multicolumn{2}{c}{Training Stage} & \multicolumn{2}{c}{World Interface} & \multicolumn{4}{c}{MindCube-Tiny} \\
\cmidrule(l{0pt}r{0pt}){3-4} \cmidrule(l{0pt}r{0pt}){5-6} \cmidrule(l{0pt}r{0pt}){7-10}
& & SFT & RL & Visual & Linguistic & Overall & Among & Rotation & Around \\
\midrule
0 & & & & & & 34.5 & 32.7 & 35.0 & 38.4 \\
1 & & & & \checkmark & & 38.5 & 36.7 & 34.0 & 46.4 \\
2 & & & & & \checkmark & 41.4 & 40.2 & 35.0 & 49.6 \\
3 & \multirow{-4}{*}{Training-free} & & & \checkmark & \checkmark & 43.8 & 43.7 & 35.0 & 51.2 \\
\midrule
4 & & \checkmark & & & & 65.2 & 66.8 & 57.0 & 68.0 \\
5 & & \checkmark & & \checkmark & \checkmark & 71.3 & 77.8 & 43.0 & 78.4 \\
\addlinespace
6 & & & \checkmark & & & 69.5 & 73.7 & 58.5 & 68.4 \\
7 & & & \checkmark & \checkmark & & 78.8 & 81.7 & 75.0 & 74.8 \\
8 & & & \checkmark & \checkmark & \checkmark & 83.9 & 84.5 & 74.0 & 90.4 \\
\addlinespace
9 & & \checkmark & \checkmark & & & 74.8 & 76.8 & 69.0 & 74.4 \\
10 & & \checkmark & \checkmark & \checkmark & & 83.9 & 85.2 & 74.5 & 88.4 \\
11 & & \checkmark & w/o SR & \checkmark & \checkmark & 89.0 & 90.0 & 83.5 & 91.2 \\
12 & \multirow{-11}{*}{Training-based} & \checkmark & \checkmark & \checkmark & \checkmark & 90.8 & 91.3 & 87.5 & 92.0 \\
\bottomrule
\end{tabular}
\end{table}

In Table~\ref{tab: ablation}, we conduct ablation studies on MindCube-Tiny with Qwen2.5-VL-7B~\cite{qwen2.5vl} as the base model (row 0). The results identify two factors underlying the performance gains. First, the world interface improves performance across different training stages. In the training-free setting, introducing both visual and linguistic interfaces improves the overall accuracy from 34.5\% to 43.8\% (rows 0-3). In the training-based setting, the full interface consistently improves overall accuracy across different training configurations, yielding gains of 6.1 points in the SFT-only configuration (rows 4-5), 14.4 points in the RL-only configuration (rows 6-8), and 16.0 points in the two-stage SFT-then-RL configuration (rows 9-12). These results highlight the importance of active world-centric spatial reasoning enabled by the world interface. \par

Second, RL more effectively elicits active exploration than SFT. Without the world interface, RL-only achieves 69.5\% overall accuracy, outperforming SFT-only at 65.2\% (rows 6 vs. 4). With the full interface, RL-only further reaches 83.9\%, substantially higher than SFT-only at 71.3\% (rows 8 vs. 5). Nevertheless, the best performance is achieved by SFT-then-RL with the full world interface, reaching 90.8\% overall accuracy (row 12). This suggests that SFT provides a useful imitation-learning initialization, while RL further optimizes the exploration policy, allowing the model to better exploit the world interface for active world-centric spatial reasoning. We further ablate the step-wise reward (SR) in RL (row 11), demonstrating the effectiveness of fine-grained reward signals. \par

More ablation studies across different base models and different 3D foundation models used for 3D world construction, as well as efficiency analyses, are provided in the appendix. \par
\section{Conclusion}
\label{sec:conclusion}

In this work, we propose View-to-World (V2W), an active exploration framework that enables VLMs to explore explicit 3D worlds for spatial reasoning. Unlike existing VLMs that passively perceive fixed-view inputs in a view-centric manner, V2W allows models to actively acquire and reason over 3D information from the underlying world, a paradigm that we refer to as active world-centric spatial reasoning. V2W constructs a VLM-compatible 3D world from sparse input views and exposes its geometry, camera poses, and language-grounded semantics through visual and linguistic interfaces, enabling VLMs to progressively gather world-grounded evidence through multi-turn interaction. Experiments show that V2W improves VLMs in a training-free setting, validating the effectiveness of active exploration over explicit 3D worlds. By further optimizing the exploration policy with reinforcement learning, V2W learns more effective exploration strategies and achieves state-of-the-art performance on spatial mental modeling benchmarks. These results highlight active world-centric spatial reasoning as a promising direction for advancing the spatial intelligence of VLMs. \par

\bibliographystyle{abbrvnat}
\bibliography{ref}

\newpage
\appendix
\section{Appendix}

\subsection{Ablation Study Across Different Base Models}

\begin{table}[h]
\centering
\footnotesize
\caption{Ablation studies across different base models on MindCube-Tiny.}
\label{tab: ablation_base}
\begin{tabular}{c@{\quad}c@{\quad}cccccc}
\toprule
\multirow{2}{*}{\#} & \multirow{2}{*}{Base Model} & \multicolumn{2}{c}{World Interface} & \multicolumn{4}{c}{MindCube-Tiny} \\
\cmidrule(l{0pt}r{0pt}){3-8}
& & Visual & Linguistic & Overall & Among & Rotation & Around \\
\midrule
0 & & & & 34.5 & 32.7 & 35.0 & 38.4 \\
1 & & \checkmark & & 38.5 & 36.7 & 34.0 & 46.4 \\
2 & & & \checkmark & 41.4 & 40.2 & 35.0 & 49.6 \\
3 & \multirow{-4}{*}{Qwen2.5-VL-7B~\cite{qwen2.5vl}} & \checkmark & \checkmark & 43.8 & 43.7 & 35.0 & 51.2 \\
\midrule
4 & & & & 30.6 & 29.2 & 31.5 & 33.2 \\
5 & & \checkmark & & 39.0 & 41.8 & 34.0 & 36.0 \\
6 & & & \checkmark & 46.8 & 44.0 & 36.5 & 61.6 \\
7 & \multirow{-4}{*}{Qwen3-VL-8B~\cite{qwen3vl}} & \checkmark & \checkmark & 50.2 & 44.8 & 38.5 & 72.4 \\
\midrule
8 & & & & 42.3 & 38.8 & 42.5 & 50.4 \\
9 & & \checkmark & & 51.0 & 56.5 & 41.0 & 45.6 \\
10 & & & \checkmark & 54.8 & 54.0 & 42.0 & 66.8 \\
11 & \multirow{-4}{*}{Qwen3-VL-235B~\cite{qwen3vl}} & \checkmark & \checkmark & 59.6 & 60.2 & 49.5 & 66.4 \\
\bottomrule
\end{tabular}
\end{table}

To verify the effectiveness of our method across different model variants and model scales, we conduct ablation studies on MindCube-Tiny~\cite{mindcube} under the training-free setting using Qwen2.5-VL-7B-Instruct~\cite{qwen2.5vl}, Qwen3-VL-8B-Instruct~\cite{qwen3vl}, and Qwen3-VL-235B-A22B-Instruct~\cite{qwen3vl} as base models, respectively. As shown in Table~\ref{tab: ablation_base}, both the visual and linguistic world interfaces consistently improve the overall performance across all base models. Specifically, the full V2W improves the overall accuracy from 34.5\% to 43.8\% on Qwen2.5-VL-7B-Instruct (row 0 vs. 3), from 30.6\% to 50.2\% on Qwen3-VL-8B-Instruct (row 4 vs. 7), and from 42.3\% to 59.6\% on Qwen3-VL-235B-A22B-Instruct (row 8 vs. 11). Notably, the Qwen3-VL series exhibits larger improvements than Qwen2.5-VL-7B, which may be attributed to its stronger tool-use capability that enables more effective utilization of the proposed interfaces for active exploration over 3D worlds. The overall performance generally improves with model scale, with Qwen3-VL-235B-A22B-Instruct achieving the best result of 59.6\% when both interfaces are enabled (row 11). In addition, the visual and linguistic interfaces exhibit consistent complementarity across all base models, with the full configuration outperforming either individual interface for all base models (row 3 vs. rows 1–2, row 7 vs. rows 5–6, and row 11 vs. rows 9–10). This suggests that the visual and textual modalities provide complementary signals during active exploration, enabling the model to capture richer 3D spatial information and thereby achieve better overall performance. \par

\subsection{Ablation Study Across Different 3D Foundation Models}

\begin{table}[h]
\centering
\footnotesize
\caption{Ablation studies across different 3D foundation models on MindCube-Tiny.}
\label{tab: vggt}
\begin{tabular}{lcccc}
\toprule
Method & \textbf{Overall} & Among & Rotation & Around \\
\midrule
V2W(Qwen2.5-VL-7B) w Depth Anything 3~\cite{da3} & 43.8 & 43.7 & 35.0 & 51.2 \\
V2W(Qwen2.5-VL-7B) w VGGT~\cite{vggt} & 42.6 & 42.2 & 35.0 & 49.6 \\
V2W(Qwen3-VL-8B) w Depth Anything 3~\cite{da3} & 50.2 & 44.8 & 38.5 & 72.4 \\
V2W(Qwen3-VL-8B) w VGGT~\cite{vggt} & 50.0 & 45.8 & 35.5 & 71.6 \\
\bottomrule
\end{tabular}
\end{table}

Our V2W leverages an off-the-shelf 3D foundation model during the 3D world construction process, with Depth Anything 3 used by default. To evaluate the robustness of V2W to different 3D foundation models and verify that its effectiveness does not depend on a specific 3D foundation model, we conduct an ablation study on MindCube-Tiny under the training-free setting by adopting VGGT, another commonly used 3D foundation model. We evaluate V2W with both Qwen2.5-VL-7B-Instruct~\cite{qwen2.5vl} and Qwen3-VL-8B-Instruct~\cite{qwen3vl} as base models, covering VLMs with different exploration capabilities and interaction behaviors during inference. As shown in Table~\ref{tab: vggt}, adopting VGGT leads to only a slight overall performance drop compared with the default Depth Anything 3 across both base models. Despite this, V2W with VGGT still substantially outperforms the corresponding base models reported in Table~\ref{tab: mindcube}. These results demonstrate that V2W exhibits robustness across different 3D foundation models, suggesting that the effectiveness of our framework does not depend on any particular 3D foundation model. \par

\subsection{Efficiency Analysis}

\begin{table}[h]
\centering
\footnotesize
\caption{Runtime efficiency analysis of the proposed pipeline on MindCube-Tiny.}
\label{tab: runtime}
\begin{tabular}{c@{\quad}ccccc}
\toprule
 & \multirow{2}{*}{3D World Construction} & \multicolumn{2}{c}{Interface Response} & \multirow{2}{*}{Model Response} & \multirow{2}{*}{Total Time} \\
\cmidrule(l{0pt}r{0pt}){3-4}
 & & Visual & Linguistic & & \\
\midrule
Avg. Time & 4.64s & 0.33s & 0.02s & 1.94s & ~6.93s / sample \\
\bottomrule
\end{tabular}
\end{table}

We further analyze the inference efficiency of the training-based V2W (Qwen2.5-VL-7B) on MindCube-Tiny. As shown in Table~\ref{tab: runtime}, the overall average inference time is approximately 6.93 seconds per sample. The main computational cost comes from 3D world construction, which takes about 5 seconds, while the interface responses introduce only negligible overhead. The model response itself takes less than 2 seconds, indicating that V2W does not rely on lengthy free-form reasoning during inference. Considering that V2W substantially outperforms models of comparable scale in spatial reasoning performance, such a runtime is acceptable. Moreover, free-form reasoning in larger models may introduce substantial inference overhead. For instance, Qwen3-VL-235B~\cite{qwen3vl} with CoT in Table~\ref{tab: mindcube} can require a thinking time comparable to, or even longer than, the total inference time of our entire pipeline, while still achieving lower performance. Rather than relying on lengthy free-form reasoning, V2W learns a more effective active exploration strategy through reinforcement learning, reducing redundant reasoning steps and unnecessary deliberation. Notably, we do not introduce any reward related to reasoning length, since we empirically find that longer reasoning does not necessarily improve the learned exploration strategy or the final answer accuracy. Instead of relying on long reasoning traces, V2W improves performance through more efficient exploration over the constructed 3D world, thereby achieving stronger spatial reasoning ability with acceptable inference efficiency. \par

\begin{figure}[t]
    \centering
    \includegraphics[width=0.99\textwidth]{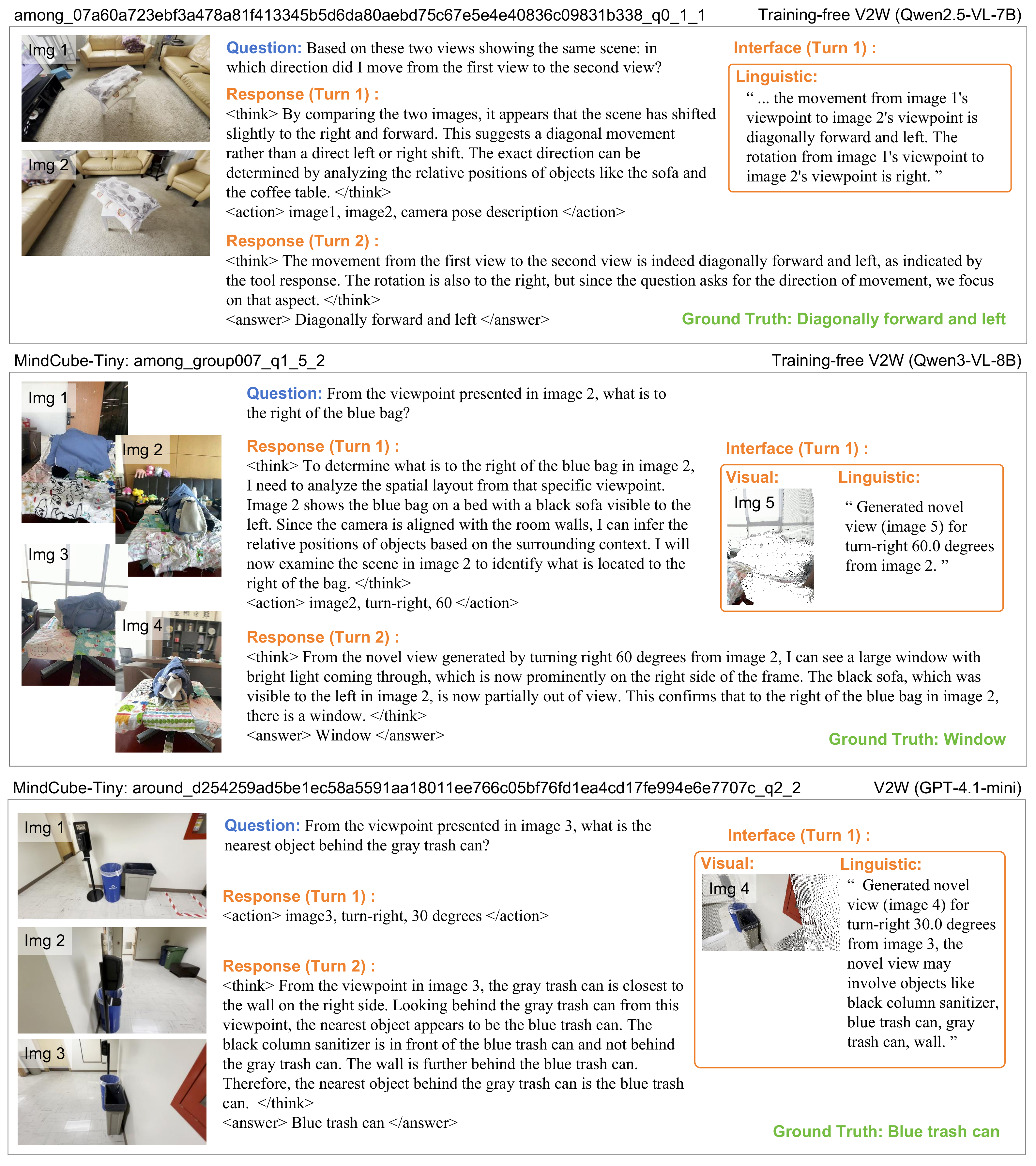}
    \caption{Additional qualitative results on MindCube-Tiny of training-free V2W with different base models.}
    \label{fig: case_free}
\end{figure}

\begin{figure}[t]
    \centering
    \includegraphics[width=0.99\textwidth]{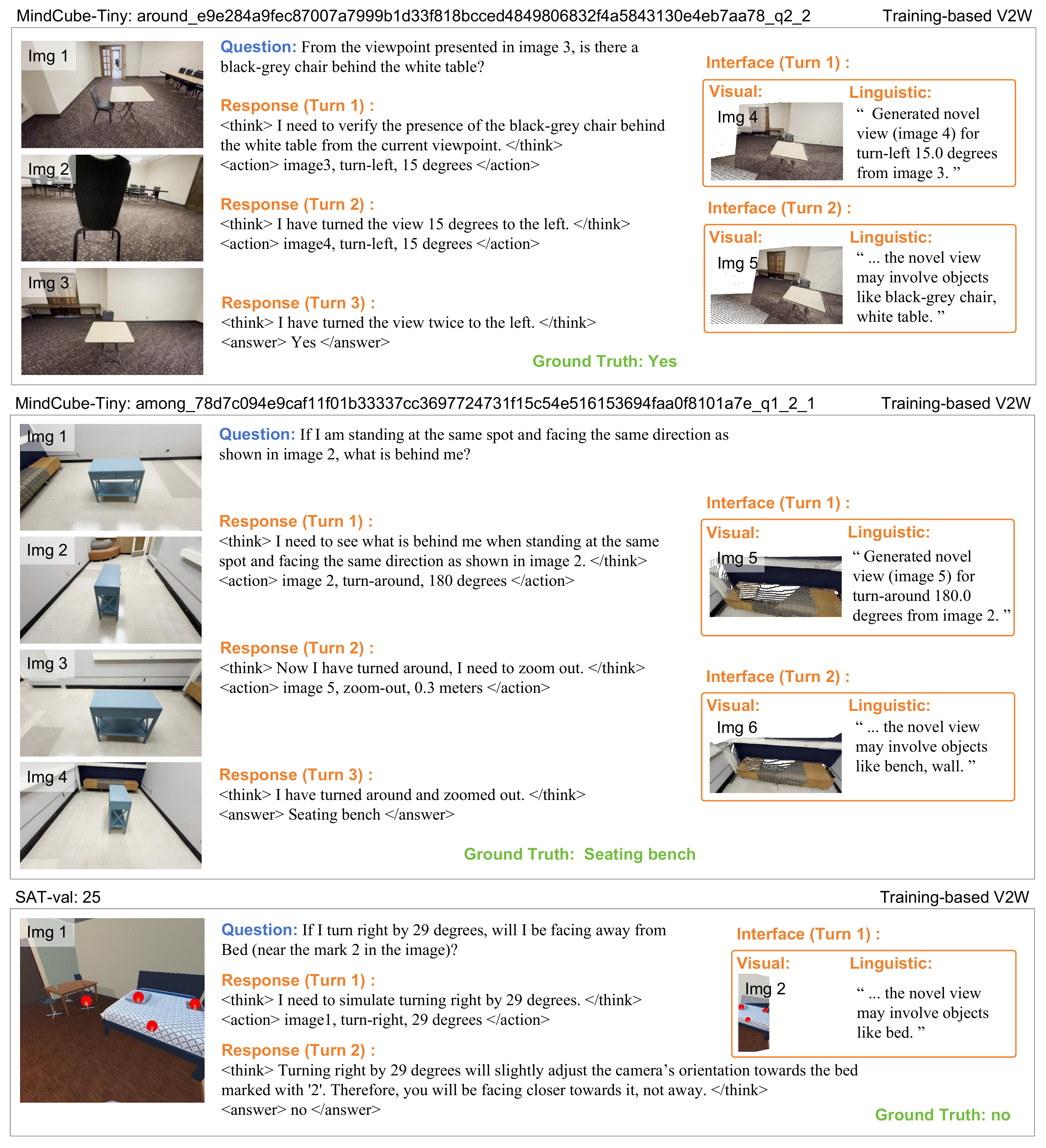}
    \caption{Additional qualitative results on MindCube-Tiny and SAT-val of training-based V2W.}
    \label{fig: case_training}
\end{figure}

\subsection{Additional Qualitative Results}

We provide additional qualitative results to further demonstrate the flexibility and effectiveness of V2W. As shown in Fig.~\ref{fig: case_free}, training-free V2W can be readily adapted to different base models, such as Qwen2.5-VL-7B~\cite{qwen2.5vl}, Qwen3-VL-8B~\cite{qwen3vl}, and GPT-4.1-mini~\cite{gpt41}, enabling them to actively explore 3D worlds without additional training. However, these models often require relatively long reasoning chains before or after exploration to determine the next action or integrate information obtained through the interface. In contrast, Fig.~\ref{fig: case_training} shows that training-based V2W performs efficient multi-round exploration in the constructed 3D worlds with minimal redundant reasoning. When the query is explicit, it can accurately identify informative exploration actions without unnecessary steps. We attribute this efficient exploration strategy to reinforcement learning, which encourages the model to learn a more concise and goal-oriented exploration policy.

\subsection{Prompt for Active Exploration via Function Calling}

\begin{figure}[t]
    \centering
    \includegraphics[width=0.99\textwidth]{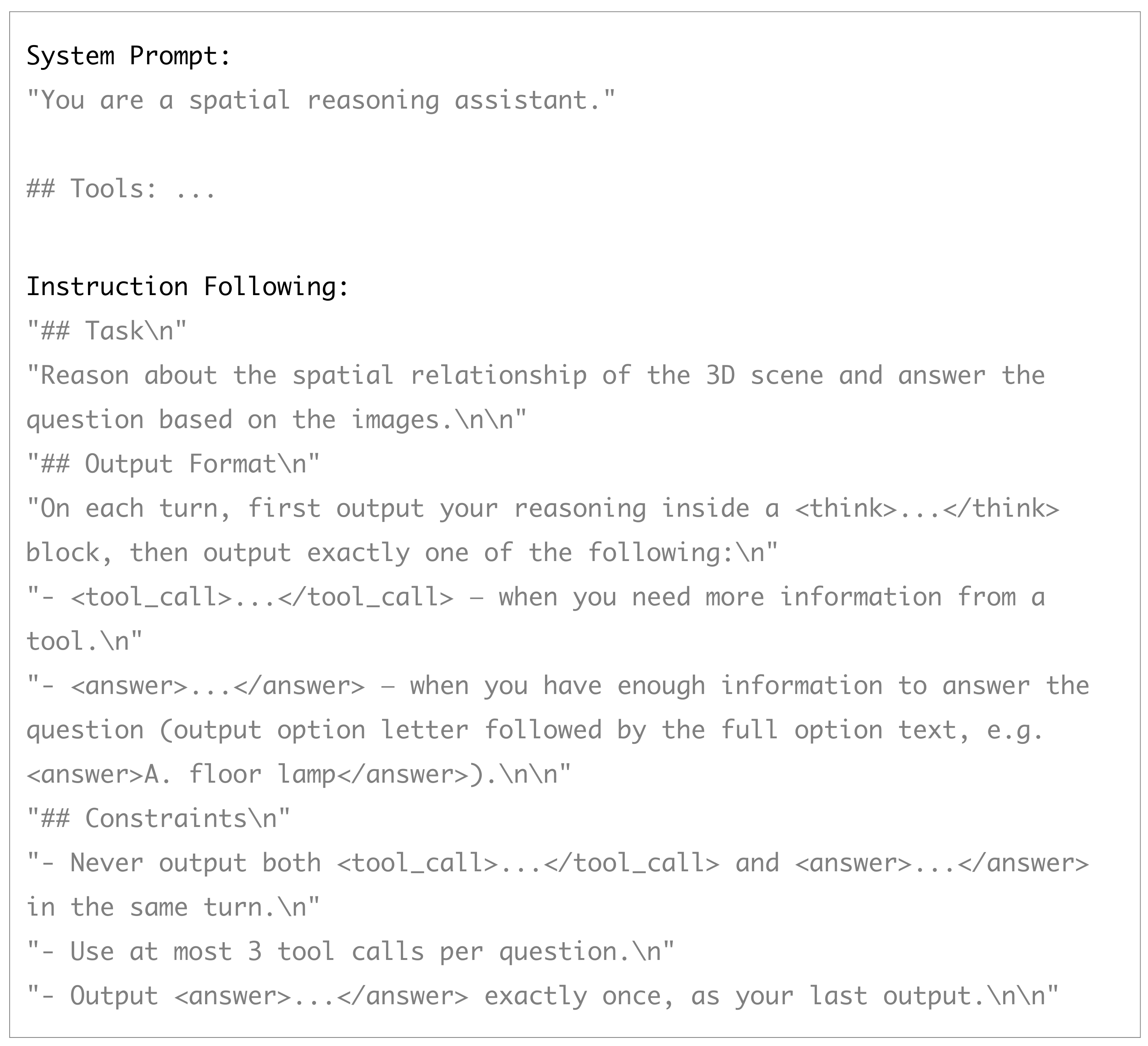}
    \caption{Example prompt of training-free and training-based V2W.}
    \label{fig: prompt}
\end{figure}

We provide an example prompt for implementing both training-free and training-based V2W in Fig.~\ref{fig: prompt}. Following a tool-usage-like paradigm, V2W enables VLMs to perform multi-turn exploration via function calling. \par

\subsection{Limitations and Future Work}

Our framework leverages off-the-shelf foundation models during 3D world construction. Although current 3D foundation models and segmentation models show strong generalization across diverse scenes, they may still fail under challenging conditions, such as low lighting or reflective objects, leading to inaccurate reconstruction or segmentation and potentially misleading subsequent reasoning. One possible direction is to introduce an additional VLM-based verifier to assess the correctness of the constructed 3D world, but this would increase the overall complexity of the framework, which we leave for future work. In addition, due to the available computational budget, we train V2W on a focused set of datasets to validate the effectiveness of reinforcement learning. Future work may scale training to broader spatial reasoning data and further study the scaling behavior of active exploration. \par

\subsection{Broader Impacts}

This work aims to improve the spatial reasoning capability of vision-language models by enabling more active interaction with 3D environments. We expect the proposed framework to benefit applications such as embodied AI, robotics, and assistive spatial understanding, while its use should follow standard privacy and safety considerations when applied to real-world visual data~\cite{towards}. \par

\subsubsection{The Use of Large Language Models (LLMs)}

In this paper, LLMs were used only as an auxiliary tool for language polishing, including improvements to grammar, clarity, and readability. They were not involved in ideation, data analysis, experimental design, or interpretation of the results.

\label{sec:appendix}

\clearpage
\newpage
\section*{NeurIPS Paper Checklist}

\begin{enumerate}

\item {\bf Claims}
    \item[] Question: Do the main claims made in the abstract and introduction accurately reflect the paper's contributions and scope?
    \item[] Answer: \answerYes{} 
    \item[] Justification: The abstract and introduction accurately reflect the contribution of studying the active exploration of VLMs and proposing a novel framework, as well as the scope of this paper. 

\item {\bf Limitations}
    \item[] Question: Does the paper discuss the limitations of the work performed by the authors?
    \item[] Answer: \answerYes{} 
    \item[] Justification: We discuss the limitations in the appendix, as well as the analysis of failure cases.

\item {\bf Theory assumptions and proofs}
    \item[] Question: For each theoretical result, does the paper provide the full set of assumptions and a complete (and correct) proof?
    \item[] Answer: \answerNA{} 
    \item[] Justification: This paper does not include theoretical results.

    \item {\bf Experimental result reproducibility}
    \item[] Question: Does the paper fully disclose all the information needed to reproduce the main experimental results of the paper to the extent that it affects the main claims and/or conclusions of the paper (regardless of whether the code and data are provided or not)?
    \item[] Answer: \answerYes{} 
    \item[] Justification: The implementation details in the Section of Experiments provide the details to reproduce the main experimental results.

\item {\bf Open access to data and code}
    \item[] Question: Does the paper provide open access to the data and code, with sufficient instructions to faithfully reproduce the main experimental results, as described in supplemental material?
    \item[] Answer: \answerYes{} 
    \item[] Justification: This paper uses public data, and the code will be released upon acceptance.

\item {\bf Experimental setting/details}
    \item[] Question: Does the paper specify all the training and test details (e.g., data splits, hyperparameters, how they were chosen, type of optimizer) necessary to understand the results?
    \item[] Answer: \answerYes{} 
    \item[] Justification: We specify all the training and test details in the Section of Experiments.

\item {\bf Experiment statistical significance}
    \item[] Question: Does the paper report error bars suitably and correctly defined or other appropriate information about the statistical significance of the experiments?
    \item[] Answer: \answerNo{} 
    \item[] Justification: Our work proposes a new framework for enhancing the active exploration capability of existing VLMs, which in turn improves their spatial reasoning ability. We validate the effectiveness of the framework on public spatial reasoning benchmarks and report results across different base models and datasets. However, due to the substantial experimental cost of VLM training and evaluation, and following common practice in recent VLM research, we do not report multi-seed results, error bars, or formal statistical significance tests.

\item {\bf Experiments compute resources}
    \item[] Question: For each experiment, does the paper provide sufficient information on the computer resources (type of compute workers, memory, time of execution) needed to reproduce the experiments?
    \item[] Answer: \answerYes{} 
    \item[] Justification: We provide the GPU resources needed in the Implementation Details of Experiments Section.
    
\item {\bf Code of ethics}
    \item[] Question: Does the research conducted in the paper conform, in every respect, with the NeurIPS Code of Ethics \url{https://neurips.cc/public/EthicsGuidelines}?
    \item[] Answer: \answerYes{} 
    \item[] Justification: This paper complies fully with the NeurIPS Code of Ethics.

\item {\bf Broader impacts}
    \item[] Question: Does the paper discuss both potential positive societal impacts and negative societal impacts of the work performed?
    \item[] Answer: \answerYes{} 
    \item[] Justification: We discuss broader impacts of this work in the appendix.
    
\item {\bf Safeguards}
    \item[] Question: Does the paper describe safeguards that have been put in place for responsible release of data or models that have a high risk for misuse (e.g., pre-trained language models, image generators, or scraped datasets)?
    \item[] Answer: \answerNA{} 
    \item[] Justification: This paper poses no such risks for misuse.

\item {\bf Licenses for existing assets}
    \item[] Question: Are the creators or original owners of assets (e.g., code, data, models), used in the paper, properly credited and are the license and terms of use explicitly mentioned and properly respected?
    \item[] Answer: \answerYes{} 
    \item[] Justification: All these assets used in this paper are correctly cited.

\item {\bf New assets}
    \item[] Question: Are new assets introduced in the paper well documented and is the documentation provided alongside the assets?
    \item[] Answer: \answerNA{} 
    \item[] Justification: This paper does not release new assets.

\item {\bf Crowdsourcing and research with human subjects}
    \item[] Question: For crowdsourcing experiments and research with human subjects, does the paper include the full text of instructions given to participants and screenshots, if applicable, as well as details about compensation (if any)? 
    \item[] Answer: \answerNA{} 
    \item[] Justification: This paper does not involve crowdsourcing nor research with human subjects.

\item {\bf Institutional review board (IRB) approvals or equivalent for research with human subjects}
    \item[] Question: Does the paper describe potential risks incurred by study participants, whether such risks were disclosed to the subjects, and whether Institutional Review Board (IRB) approvals (or an equivalent approval/review based on the requirements of your country or institution) were obtained?
    \item[] Answer: \answerNA{} 
    \item[] Justification: This paper does not involve crowdsourcing nor research with human subjects.

\item {\bf Declaration of LLM usage}
    \item[] Question: Does the paper describe the usage of LLMs if it is an important, original, or non-standard component of the core methods in this research? Note that if the LLM is used only for writing, editing, or formatting purposes and does \emph{not} impact the core methodology, scientific rigor, or originality of the research, declaration is not required.
    \item[] Answer: \answerYes{} 
    \item[] Justification: Our method is based on vision-language models (VLMs), and the experiments use existing VLMs as base models for both training-free evaluation and post-training. In addition, we use Qwen3-VL-235B for training data processing. These uses are part of the research methodology and experimental pipeline, and are described in the paper. We also used LLMs for translation, grammar checking, and writing assistance only; such usage did not affect the core methodology, scientific rigor, or originality of the research.

\end{enumerate}

\end{document}